\documentclass{article}

\usepackage{PRIMEarxiv}

\usepackage[utf8]{inputenc} %
\usepackage[T1]{fontenc}    %
\usepackage{hyperref}       %
\usepackage{url}            %
\usepackage{booktabs}       %
\usepackage{amsfonts}       %
\usepackage{nicefrac}       %
\usepackage{microtype}      %
\usepackage{lipsum}
\usepackage{fancyhdr}       %
\usepackage{graphicx}       %
\usepackage{xcolor}
\graphicspath{{media/}}     %
\usepackage{amsmath}
\usepackage{comment}
\usepackage{algorithm}
\usepackage{algorithmic}

\newcommand{\uterusLowIoU}{+5.15}
\newcommand{\uterusLowDice}{+3.13}
\newcommand{\umdLowIoU}{+6.76}
\newcommand{\umdLowDice}{+5.70}
\newcommand{\uterusLowHD}{+7.03}
\newcommand{\umdLowHD}{+18.12}

\newcommand{\uterusHighIoU}{-5.52}
\newcommand{\uterusHighDice}{-3.76}
\newcommand{\uterusHighHD}{-6.95}
\newcommand{\umdHighIoU}{-9.88}
\newcommand{\umdHighDice}{-9.04}
\newcommand{\umdHighHD}{-9.33}

\newcommand{\hepGammaMIoU}{-0.13}
\newcommand{\hepGammaAIoU}{+0.02}
\newcommand{\hepGammaMDice}{-0.01}
\newcommand{\hepGammaADice}{+0.22}
\newcommand{\hepGammaHD}{-0.23}

\newcommand{\hepContrastMIoU}{-0.24}
\newcommand{\hepContrastAIoU}{+0.11}
\newcommand{\hepContrastMDice}{-0.09}
\newcommand{\hepContrastADice}{+0.39}
\newcommand{\hepContrastHD}{-1.66}

\newcommand{\hepBlurMIoU}{+2.25}
\newcommand{\hepBlurAIoU}{+2.72}
\newcommand{\hepBlurMDice}{+2.05}
\newcommand{\hepBlurADice}{+1.96}
\newcommand{\hepBlurHD}{+5.59}

\newcommand{\hepNoiseMIoU}{+3.19}
\newcommand{\hepNoiseAIoU}{+2.91}
\newcommand{\hepNoiseMDice}{+2.65}
\newcommand{\hepNoiseADice}{+1.77}
\newcommand{\hepNoiseHD}{+1.71}

\newcommand{\hepVoteMIoU}{+1.60}
\newcommand{\hepVoteAIoU}{+1.94}
\newcommand{\hepVoteMDice}{+1.42}
\newcommand{\hepVoteADice}{+1.44}
\newcommand{\hepVoteHD}{-1.98}

\newcommand{\uterusRmGammaIoU}{-4.69}
\newcommand{\uterusRmGammaDice}{-3.12}
\newcommand{\uterusRmGammaHD}{-8.53}
\newcommand{\umdRmGammaIoU}{-1.15}
\newcommand{\umdRmGammaDice}{-1.02}
\newcommand{\umdRmGammaHD}{-2.96}

\newcommand{\uterusRmContrastIoU}{-4.59}
\newcommand{\uterusRmContrastDice}{-3.08}
\newcommand{\uterusRmContrastHD}{-8.61}
\newcommand{\umdRmContrastIoU}{-1.47}
\newcommand{\umdRmContrastDice}{-1.26}
\newcommand{\umdRmContrastHD}{-3.27}

\newcommand{\uterusRmBlurIoU}{-2.54}
\newcommand{\uterusRmBlurDice}{-1.68}
\newcommand{\uterusRmBlurHD}{-3.54}
\newcommand{\umdRmBlurIoU}{-2.67}
\newcommand{\umdRmBlurDice}{-2.61}
\newcommand{\umdRmBlurHD}{-3.99}

\newcommand{\uterusRmNoiseIoU}{-4.22}
\newcommand{\uterusRmNoiseDice}{-2.80}
\newcommand{\uterusRmNoiseHD}{-7.64}
\newcommand{\umdRmNoiseIoU}{-5.95}
\newcommand{\umdRmNoiseDice}{-4.97}
\newcommand{\umdRmNoiseHD}{-8.60}

\newcommand{\uterusVoteIoUDelta}{+0.42}
\newcommand{\uterusVoteDiceDelta}{+0.34}
\newcommand{\uterusVoteHDDelta}{-1.78}
\newcommand{\umdVoteIoUDelta}{+0.49}
\newcommand{\umdVoteDiceDelta}{+0.73}
\newcommand{\umdVoteHDDelta}{-2.73}

\title{SegTTA: Training-Free Test-Time Augmentation for Zero-Shot Medical Imaging Segmentation
}

\author{
    \textbf{Yihong Yao}$^{1*}$ ~
    \textbf{Chunlei Li}$^{2*}$ ~
    \textbf{Canxuan Gang}$^{1*}$ ~
    \textbf{Wenzhi Hu}$^{1*}$ ~
    \textbf{Zeyu Zhang}$^{1\dag}$ ~
    \textbf{Hao Zhang}$^{3}$ ~
    \textbf{Xiaoyan Li}$^{2\ddag}$ \vspace{0.1cm}\\
    $^{1}$AI Geeks \quad
    $^{2}$Qingdao Municipal Hospital \quad
    $^{3}$University of Chinese Academy of Sciences\vspace{0.06cm}\\
    \small $^*$Equal contribution. $^\dag$Project lead.
    $^\ddag$Corresponding author: xiaoyanli.qmh.offical@gmail.com.
}

\begin{document}
\maketitle

\begin{abstract}
Increasingly advanced data augmentation techniques have greatly aided clinical medical research, increasing data diversity, and improving model generalization capabilities. Although most current basic models exhibit strong generalization abilities, image quality varies due to differences in equipment and operators. To address these challenges, we present \textbf{SegTTA}, a framework that improves medical image segmentation without model retraining by combining four augmentations (Gamma correction, Contrast enhancement, Gaussian blur, Gaussian noise) with weighted voting across multiple MedSAM2 checkpoints. Experiments demonstrate consistent improvements across three diverse datasets: healthy uterus segmentation, uterine myoma detection, and multi-class hepatic structure segmentation. Ablation studies reveal that large organs benefit from intensity augmentations while small lesions require noise augmentations. The voting threshold controls the coverage-precision trade-off, enabling task-specific optimization for different clinical requirements. Ultimately, on a multiclass hepatic vessel dataset, compared to MedSAM2 baselines, our method achieves an increase of \textbf{1.6} in mIoU and \textbf{1.9} in aIoU, along with a reduction of approximately \textbf{2.0} in HD95.
Code will be available at \url{https://github.com/AIGeeksGroup/SegTTA}.
\end{abstract}


\section{Introduction}

Medical image analysis, encompassing tasks such as segmentation, detection, and classification, is a cornerstone of modern clinical decision support systems \cite{goceri2023medical}. To enhance the generalization ability and reliability of models, data augmentation techniques have emerged\cite{lopes2019improving}. Data augmentation is a cost-effective and powerful strategy that artificially increases the diversity of data distribution, thereby significantly improving the effective size and diversity of available datasets without collecting new patient scan data. In medical image analysis, data augmentation not only prevents overfitting but is also essential for creating models that can adapt to the highly variability of clinical environments\cite{qi2025mediaug}.

While foundation models like MedSAM2 show strong generalization\cite{ma2025medsam2}, performance gaps remain in clinical deployment, particularly for ultrasound imaging where quality varies significantly across operators and equipment. Test-time augmentation (TTA) aggregates predictions from augmented test images to improve robustness without retraining\cite{ma2024test}. However, standard TTA strategies developed for natural images may not address medical imaging needs where subtle intensity variations and precise boundary delineation are critical for clinical utility\cite{nazzal2024improving}.

To address these challenges, we propose SegTTA, a TTA framework tailored for medical segmentation that incorporates medical-specific augmentations and adaptive voting strategies. Our framework applies four complementary augmentations, Gamma correction, Contrast enhancement, Gaussian blur, and Gaussian noise, which target common clinical variations~\cite{kallel2018ct}. We then combine predictions through weighted voting with adjustable thresholds~\cite{tasci2021voting}. By leveraging multiple MedSAM2 checkpoints trained on diverse modalities, we create robust ensemble predictions particularly effective for challenging tasks like small lesion segmentation and multi-class segmentation such as hepatic vessels and tumors.

The framework's adaptability enables optimization for various clinical priorities. The main contributions of this work are summarized as follows:

\begin{itemize}
    \item \textbf{SegTTA}, a \textit{training-free} framework, is presented to enhance MedSAM2 robustness through four medical-specific augmentations, requiring no parameter updates.
    
    \item An adaptive \textit{weighted voting algorithm} is introduced to aggregate predictions, utilizing adjustable thresholds to balance segmentation coverage and boundary precision.
    
    \item Experiments on three diverse datasets (UterUS, UMD, HepaticVessel) demonstrate consistent performance gains, achieving average improvements of \textbf{0.42\% IoU}, \textbf{0.49\% IoU}, and \textbf{1.60\% mIoU}, respectively.
\end{itemize}

\section{Related Work}
\label{sec:headings}

Segmentation research has progressed with datasets and architectures such as BHSD, Segstitch, and Thin-Thick Adapter \cite{wu2023bhsd,tan2024segstitch,zhang2023thin}, while ESA, DOEI, GAMED-Snake, SegKAN, MARL-MambaContour, Unified Snake, and SSS further expand model design \cite{ge2025esa,zhu2025doei,zhang2025gamed,tan2024segkan,zhang2025marl,zhang2025unified,zhu2025sss}. Training-free augmentation has also been pursued through MedSAMix \cite{yang2025medsamix}. Detection has advanced with MSDet and MedDet \cite{cai2024msdet,zhang2024meddet}, complemented by PedDet, EPDD-YOLO, and surveys on lung cancer detection \cite{zhao2025peddet,luo2025epdd,cai2025medical}. Representation learning has evolved through multimodal and long-tailed modeling, including MMCLIP and JointViT \cite{wu2024mmclip,zhang2024jointvit}, low-rank matrix learning \cite{ji2024efficient}, MedConv \cite{qi2025medconv}, and pathology-driven survival prediction with PathoHR \cite{luo2025pathohr}. Diagnostic applications include diabetes detection, fracture instability prediction, prostate cancer analysis, and traumatic brain injury assessment \cite{zhang2024deep,zhao2024landmark,qi2025projectedex,hiwase2025can}. Building on these developments, the present work addresses uterus and ovarian segmentation with MedSAM2 \cite{ma2025medsam2} in a free-training setting, using hepatic vessel data for auxiliary validation.

Test-time augmentation (TTA) has emerged as a key strategy to improve model robustness without retraining by aggregating predictions from multiple augmented views of a test image \cite{ma2024test}. Recent studies have investigated TTA in medical segmentation through random circular shifts in MedSAM \cite{nazzal2024improving}, generative diffusion-based augmentation \cite{ma2024test}, and SAM2 extensions for few-shot volumetric tasks \cite{zu2025rethinking}. However, standard TTA strategies developed for natural images often fail to address specific medical imaging needs, where subtle intensity variations and precise boundary delineation are critical for clinical utility \cite{nazzal2024improving}. Benchmarks including MediAug \cite{qi2025mediaug} and surveys on pre-trained SAM \cite{wu2025pre} demonstrate the effectiveness of such augmentation strategies, highlighting how ensemble approaches can mitigate the instability of single-model predictions to produce more coherent segmentation maps.

\begin{figure}[t]
    \centering
    \includegraphics[width=1\textwidth]{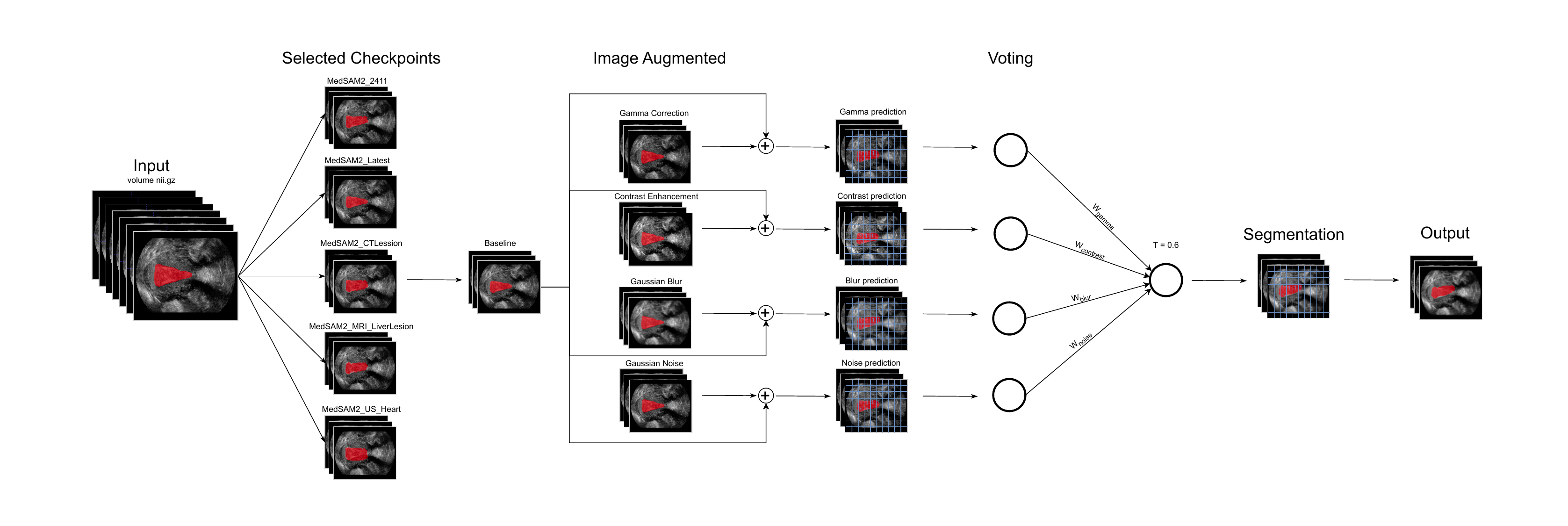}
    \caption{Framework of SegTTA. Baseline outputs from multiple MedSAM2 checkpoints and augmented predictions are fused through a voting strategy to improve segmentation robustness.}
    \label{fig:main-graph}
\end{figure}

\section{Method}
\label{sec:others}
\subsection{Overview}
Our framework employs MedSAM2 in a free-training setting, avoiding additional fine-tuning or supervised training. Multiple pretrained checkpoints of MedSAM2 are used to establish a baseline prediction. To enhance robustness at inference, we integrate a test-time augmentation scheme that generates perturbed views of the input CT scans. Each augmented image is segmented by MedSAM2 independently, and the resulting predictions are fused with the baseline output through a voting strategy. This design enables the model to better handle acquisition variability and improves consistency across different anatomical regions, as shown in Figure~\ref{fig:main-graph}.

\subsection{Visual Augmentation}

Four augmentations were selected to reflect the intrinsic variability and noise characteristics of CT and MRI imaging. \textit{Gaussian blur} was included to emulate motion artifacts and reduced resolution, which are common in dynamic acquisitions and low-quality scans. \textit{Gaussian noise injection} accounts for detector and electronic noise, particularly evident in low-dose CT and high-field MRI where signal-to-noise ratio is limited. \textit{Gamma correction} models global intensity shifts arising from scanner calibration differences and variations in tissue contrast across patients. \textit{Contrast enhancement} further captures changes in tissue-to-background separability caused by acquisition protocols or pathological conditions. Together, these augmentations mimic realistic sources of distortion and variability in medical imaging, thereby improving the robustness and generalizability of segmentation.

\subsubsection{Gaussian Blur}
Gaussian blur \cite{lopes2019improving} simulates reduced resolution and motion artifacts in CT imaging by smoothing local variations and attenuating sharp boundaries. This compels the model to capture global structural cues rather than rely on local edge sharpness. The blurred image $I'(x,y)$ is obtained by convolving the input $I(x,y)$ with a Gaussian kernel $G(i,j)$:
\[
I'(x,y) = \sum_{i=-k}^{k}\sum_{j=-k}^{k} G(i,j)\,I(x-i,y-j), \quad 
G(i,j) = \frac{1}{2\pi\sigma^2}\exp\!\left(-\frac{i^2+j^2}{2\sigma^2}\right).
\]

\subsubsection{Noise Injection}
Noise injection \cite{erichson2022noisymix} reflects stochastic perturbations introduced during CT acquisition, such as detector noise or reconstruction artifacts. It improves tolerance to background fluctuations and forces the model to ignore irrelevant texture. The perturbed image is defined as
\[
I'(x,y) = I(x,y) + \mathcal{N}(0,\sigma^2),
\]
where $\mathcal{N}(0,\sigma^2)$ denotes zero-mean Gaussian noise with variance $\sigma^2$.

\subsubsection{Gamma Correction}
Gamma correction \cite{kallel2018ct} introduces non-linear intensity transformations, simulating variations in scanner calibration and acquisition protocols. It alters brightness distributions and evaluates robustness under global intensity shifts. The operation is defined as
\[
I'(x,y) = \left(\frac{I(x,y)}{I_{\max}}\right)^\gamma I_{\max},
\]
where $I_{\max}$ is the maximum intensity and $\gamma$ controls the transformation. Values $\gamma>1$ darken the image, whereas $\gamma<1$ brighten it.

\subsubsection{Contrast Enhancement}
Contrast enhancement \cite{goceri2023medical} linearly scales image intensities, adjusting separability between tissues and background. This augmentation tests whether the model maintains stability under varying contrast conditions. The operation is expressed as
\[
I'(x,y) = \alpha I(x,y) + \beta,
\]
where $\alpha$ determines contrast level and $\beta$ shifts overall brightness. Intensities are clipped to the valid range of the image.

\subsection{Voting Algorithm}
To obtain a robust final prediction from multiple augmented inputs, we adopt a voting algorithm \cite{tasci2021voting} that aggregates the segmentation outputs of MedSAM2 under different test-time augmentations. Each augmented image is independently segmented, yielding a set of probability maps $\{P_1, P_2, \ldots, P_N\}$ corresponding to $N$ augmentations. These maps are fused by combining majority voting and confidence-weighted voting strategies.

\subsubsection{Majority Voting}
In majority voting \cite{lam1997application}, the final label $\hat{y}(x)$ for pixel $x$ is determined by the most frequent prediction among all augmentation outputs:
\[
\hat{y}(x) = \arg\max_{c} \sum_{i=1}^{N} \mathbf{1}\!\left( \arg\max_{c'} P_i(x,c') = c \right),
\]
where $c$ denotes a candidate class and $\mathbf{1}(\cdot)$ is the indicator function. This approach emphasizes consistency across augmented views and reduces the influence of outlier predictions.

\subsubsection{Confidence-Weighted Voting}
While majority voting treats all augmentations equally, confidence-weighted voting \cite{toth2008classification} exploits the probability distribution provided by MedSAM2. The aggregated decision is defined as
\[
\hat{y}(x) = \arg\max_{c} \sum_{i=1}^{N} w_i(x) \cdot P_i(x,c),
\]
where the weight $w_i(x)$ corresponds to the maximum probability at pixel $x$ for the $i$-th augmentation:
\[
w_i(x) = \max_{c} P_i(x,c).
\]
This weighting scheme assigns higher influence to confident predictions, thereby reducing the effect of uncertain outputs.

\subsubsection{Final Aggregation}
In practice, the two strategies are complementary: majority voting provides stability across perturbations, while confidence-weighted voting leverages pixel-level uncertainty to refine predictions. By combining these algorithms, the final segmentation achieves greater robustness and accuracy, with uterus and ovarian datasets serving as the primary benchmarks and hepatic vessel data included only as auxiliary validation to demonstrate feasibility and effectiveness.

\begin{algorithm}[H]
\caption{SegTTA: Test-Time Augmentation for Medical Segmentation}
\label{alg:segtta}
\begin{algorithmic}[1]
\REQUIRE Input scan $I$, MedSAM2 checkpoints $\{\text{Model}_1, ..., \text{Model}_M\}$, \\
         Augmentation set $\mathcal{T} = \{T_{\gamma}, T_{contrast}, T_{blur}, T_{noise}\}$, \\
         Voting threshold $\tau$ (default: 0.6)
\ENSURE Final segmentation mask $\hat{y}$

\STATE \textbf{Step 1: Baseline Inference}
\FOR{each checkpoint $\text{Model}_j$}
    \STATE Obtain baseline probability map: $P_j^{base} = \text{Model}_j(I)$
\ENDFOR

\STATE \textbf{Step 2: Test-Time Augmentation}
\FOR{each augmentation $T_i \in \mathcal{T}$}
    \STATE Generate perturbed view: $I'_i = T_i(I)$
    \FOR{each checkpoint $\text{Model}_j$}
        \STATE Obtain augmented probability map: $P_{i,j}^{aug} = \text{Model}_j(I'_i)$
    \ENDFOR
\ENDFOR

\STATE \textbf{Step 3: Weighted Voting Aggregation}
\STATE Collect all predictions: $\mathcal{P} = \{P_j^{base}\} \cup \{P_{i,j}^{aug}\}$
\FOR{each pixel $x$ and candidate class $c$}
    \STATE Calculate confidence-weighted score:
    \STATE \quad $S(x,c) = \sum_{P_k \in \mathcal{P}} w_k(x) \cdot P_k(x,c)$
    \STATE \quad where $w_k(x) = \max_{c'} P_k(x,c')$
    \STATE Apply threshold voting:
    \STATE \quad $\hat{y}(x) = \begin{cases} 
                  \arg\max_{c} S(x,c) & \text{if } \max_c S(x,c) \geq \tau \\
                  \text{background} & \text{otherwise}
                  \end{cases}$
\ENDFOR

\RETURN Final segmentation $\hat{y}$
\end{algorithmic}
\end{algorithm}

\begin{figure}[t]
    \centering
    \includegraphics[width=1\textwidth]{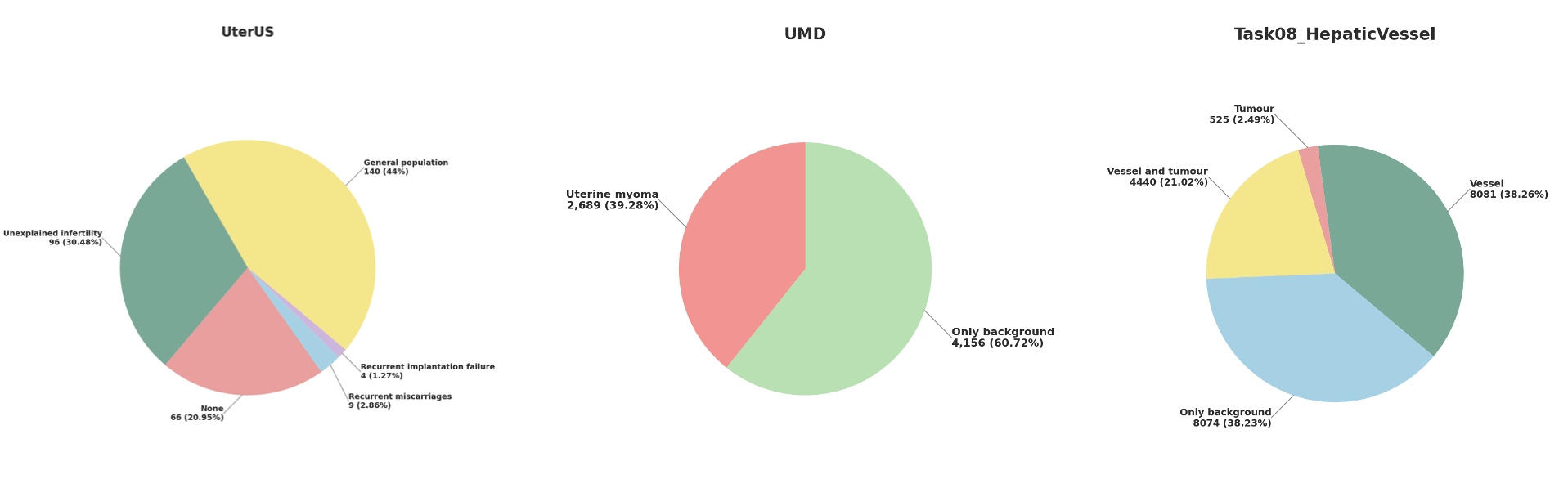}
    \caption{\small UterUS dataset \cite{bones2024uterus} with five categories (1.27\%-44\%). UMD dataset \cite{umd2024} with two categories (39.28\%-60.71\%). HepaticVessel dataset \cite{task08hepatic2019} with five categories (2.49\%-38.26\%).}
    \label{fig:dataset-pie}
\end{figure}

\section{Experiments}

\subsection{Dataset and Evaluation Metrics} 

\paragraph{Datasets}
Figure~\ref{fig:dataset-pie} shows the category distributions of UterUS, UMD, and HepaticVessel. The plots highlight class imbalance, including the predominance of general population cases in UterUS, the small fraction of myoma slices in UMD, and the heterogeneous vessel–tumor composition in HepaticVessel, providing context for evaluating segmentation robustness.

\textit{UterUS Dataset: }
The UterUS dataset \cite{bones2024uterus} is a single-class semantic segmentation resource for the endometrial cavity from 3D transvaginal ultrasound volumes. It contains 141 annotated scans in \texttt{.nii.gz} format with binary masks, while 174 unannotated volumes are excluded. Each scan includes metadata such as medical center, sample number, ultrasound machine, and clinical classification. The dataset is divided into five groups: General population (G, 140 cases, 44\%), Unexplained infertility (I, 96 cases, 30.48\%), Recurrent miscarriage (M, 9 cases, 2.86\%), Recurrent implantation failure (RIF, 4 cases, 1.27\%), and Uncategorized (66 cases, 20.95\%), providing a benchmark for uterus cavity segmentation across diverse clinical conditions.

\textit{UMD Dataset: }
The UMD dataset \cite{umd2024} contains 6,845 T2-weighted sagittal MRI slices from 300 patients, with pixel-wise annotations for uterine myomas covering nine FIGO types and hybrid forms. Each slice is labeled with four classes: uterine wall (1), uterine cavity (2), myoma (3), and nabothian cyst (4). Among slices, 39.28\% contain myomas. Only myoma-containing slices are used in this study, treating the dataset as single-class segmentation for myomas, suitable for evaluating methods on small-volume connected components.

\textit{HepaticVessel Dataset: }
The Task08\_HepaticVessel dataset \cite{task08hepatic2019} from the Medical Segmentation Decathlon is a multi-class semantic segmentation resource for 3D segmentation of hepatic vessels and liver tumors from abdominal CT scans. It contains 303 contrast-enhanced portal-venous CT volumes with vessel and tumor masks for training, and 139 unlabeled volumes for testing. Each voxel is labeled as vessel (1), tumor (2), or background (0). Among all voxels, 38.26\% are vessels, 2.49\% tumors, 21.02\% both, and 38.23\% background. This multi-class dataset supports evaluation of segmentation models on fine, tubular, and connected vascular structures in heterogeneous livers.

\paragraph{Metrics}

We follow SegReg \cite{zhang2024segreg}, which uses five metrics: agnostic IoU (aIoU), agnostic Dice (aDice), mean IoU (mIoU), mean Dice (mDice), and the 95th percentile Hausdorff Distance (HD95). aIoU and aDice measure overall region overlap without class labels. mIoU and mDice average per-class accuracy and reveal segmentation bias, with mDice more responsive to small structures. HD95 quantifies boundary error while reducing outlier impact. Together, these metrics capture region accuracy, class-level consistency, and boundary precision.

\subsection{Implementation Details}
Experiments utilized two gynecological datasets, UterUS (3D ultrasound) and UMD (T2-weighted MRI), as primary benchmarks, with the Hepatic Vessel dataset serving as auxiliary validation. Consistent with the training-free nature of our framework, no data splitting for training or fine-tuning was performed; instead, the pre-trained MedSAM2 model was applied directly to all annotated volumes for inference evaluation. To ensure reproducibility of stochastic test-time augmentations (e.g., Gaussian noise and blur), a fixed random seed of \textbf{2024} was initialized for all experiments. Computing was conducted on an NVIDIA A100 GPU (80GB) with CUDA 12.4 and an Intel Xeon CPU @ 2.20GHz.

\begin{table}[t]
\centering
\caption{Segmentation metrics comparison of MedSAM2 models on UterUS \cite{bones2024uterus} and UMD\cite{umd2024} dataset.}
\begin{tabular}{lccc|ccc}
\toprule
 & \multicolumn{3}{c}{\textbf{UterUS}} & \multicolumn{3}{c}{\textbf{UMD}} \\
\cmidrule(lr){2-4} 
\cmidrule(lr){5-7} 
 \textbf{Models} & \textbf{IoU} & \textbf{Dice} & \textbf{HD95} & \textbf{IoU} & \textbf{Dice} & \textbf{HD95} \\
\midrule
MedSAM2\_2411 & 78.67 & 87.05 & 32.12 & 78.75 & 85.64 & 21.95 \\
MedSAM2\_US\_Heart & 79.64 & 88.26 & 48.08 & 83.68 & 87.91 & 37.16 \\
MedSAM2\_MRI\_LiverLesion & 81.23 & 89.26 & 33.61 & 81.50 & 88.31 & 16.96 \\
MedSAM2\_CTLesion & 79.52 & 88.11 & 30.32 & 81.28 & 87.68 & \textbf{20.43} \\
MedSAM2\_latest & 77.85 & 86.65 & \textbf{23.56} & 79.08 & 85.60 & 24.55 \\
\midrule
\textbf{SegTTA (Ours)} & \textbf{81.65} & \textbf{89.60} & 31.83 & \textbf{84.17} & \textbf{88.64} & 34.43\\
\bottomrule
\end{tabular}
\label{tab:medsam2_UterUS_results}
\end{table}

\begin{table}[t]
\centering
\caption{Segmentation metrics comparison of MedSAM2 models on Task08\_HepaticVessel dataset \cite{task08hepatic2019}.}
\begin{tabular}{lccccc}
\toprule
\textbf{Models} & \textbf{mIoU} & \textbf{aIoU} & \textbf{mDice} & \textbf{aDice} & \textbf{HD95} \\
\midrule
MedSAM2\_2411            & 73.99 & 79.97 & 80.45 & 87.73 & 29.48 \\
MedSAM2\_US\_Heart               & 75.87 & 81.21 & 82.28 & \textbf{88.16} & 33.06 \\
MedSAM2\_MRI\_LiverLesion        & 69.53 & 76.71 & 75.98 & 85.19 & 28.03 \\
MedSAM2\_CTLesion                & 72.86 & 79.27 & 79.40 & 87.15 & 27.63 \\
MedSAM2\_latest         & 66.40 & 72.46 & 73.58 & 81.42 & \textbf{27.14} \\
\midrule
\textbf{SegTTA (Ours)} & \textbf{77.47} & \textbf{83.15} & \textbf{83.70} & \textbf{89.60} & 31.08\\
\bottomrule
\end{tabular}
\end{table}

\subsection{Main Results}

The quantitative evaluation of SegTTA across three diverse medical imaging datasets: UterUS (ultrasound), UMD (MRI), and HepaticVessel (CT), demonstrates consistent performance enhancements over the individual MedSAM2 baseline checkpoints \cite{bones2024uterus, umd2024, task08hepatic2019}. As summarized in Tables 1, 2, 3 and 4, our framework successfully improves segmentation accuracy without requiring any model retraining or fine-tuning \cite{ma2025medsam2}.

On the single-class segmentation tasks, SegTTA achieves an IoU of 81.65\% and a Dice score of 89.60\% for the UterUS dataset, surpassing the best-performing individual checkpoint \cite{bones2024uterus}. Similarly, for the UMD dataset targeting uterine myoma, the framework attains an IoU of 84.17\% and a Dice score of 88.64\%, effectively identifying challenging small lesions \cite{umd2024}. In the multi-class HepaticVessel scenario, SegTTA yields a mean IoU (mIoU) of 77.47\%, significantly outperforming the baselines in delineating complex vascular and tumor structures \cite{task08hepatic2019}.

The results further indicate that the weighted voting mechanism (threshold = 0.6) provides a robust balance between region overlap and boundary precision, leading to a consistent reduction in HD95 across most tasks \cite{tasci2021voting, toth2008classification}. These findings validate the effectiveness of the training-free ensemble approach in addressing the acquisition variability inherent in diverse clinical environments \cite{ma2025medsam2}.

\begin{table}[t]
\centering
\caption{Segmentation metrics comparison of MedSAM2\_MRI\_LiverLesion models on UterUS \cite{bones2024uterus} dataset and MedSAM2\_US\_Heart on UMD\cite{umd2024} dataset.}
\begin{tabular}{lccc|ccc}
\toprule
 & \multicolumn{3}{c}{\textbf{UterUS}} & \multicolumn{3}{c}{\textbf{UMD}} \\
\cmidrule(lr){2-4} 
\cmidrule(lr){5-7} 
 \textbf{Models} & \textbf{IoU} & \textbf{Dice} & \textbf{HD95} & \textbf{IoU} & \textbf{Dice} & \textbf{HD95} \\
\midrule
Baseline & 81.23 & 89.26 & 33.61 & 83.68 & 87.91 & 37.16 \\
Gamma correction  
         & 81.14\textsubscript{\textcolor{red}{-0.09}} 
         & 89.20\textsubscript{\textcolor{red}{-0.06}} 
         & 33.58\textsubscript{\textcolor{green}{-0.03}} 
         & 83.93\textsubscript{\textcolor{green}{+0.25}}
         & 88.09\textsubscript{\textcolor{green}{+0.18}}
         & 37.04\textsubscript{\textcolor{green}{-0.12}}\\

Contrast enhancement 
		 & 80.66\textsubscript{\textcolor{red}{-0.57}} 
         & 88.91\textsubscript{\textcolor{red}{-0.35}} 
         & 33.48\textsubscript{\textcolor{green}{-0.13}}
         & 84.51\textsubscript{\textcolor{green}{+0.83}}
         & 88.66\textsubscript{\textcolor{green}{+0.75}}
         & 37.65\textsubscript{\textcolor{red}{+0.49}}\\
Gaussian blur 
		 & 83.02\textsubscript{\textcolor{green}{+1.79}} 
         & 90.37\textsubscript{\textcolor{green}{+1.11}} 
         & 31.20\textsubscript{\textcolor{green}{-2.41}}
         & 81.61\textsubscript{\textcolor{red}{-2.07}}
         & 86.63\textsubscript{\textcolor{red}{-1.28}}
         & 38.69\textsubscript{\textcolor{red}{+1.53}}\\

Gaussian noise 
		 & 81.71\textsubscript{\textcolor{green}{+0.48}} 
         & 89.56\textsubscript{\textcolor{green}{+0.30}} 
         & 33.64\textsubscript{\textcolor{red}{+0.03}}
         & 85.07\textsubscript{\textcolor{green}{+1.39}}
         & 88.99\textsubscript{\textcolor{green}{+1.08}}
         & 44.46\textsubscript{\textcolor{red}{+7.30}}\\
\midrule
Weighted Voting (0.6)
		 & 81.65\textsubscript{\textcolor{green}{+0.42}} 
         & 89.60\textsubscript{\textcolor{green}{+0.34}} 
         & 31.83\textsubscript{\textcolor{green}{-1.78}}
         & 84.17\textsubscript{\textcolor{green}{+0.49}}
         & 88.64\textsubscript{\textcolor{green}{+0.73}}
         & 34.43\textsubscript{\textcolor{green}{-2.73}}\\
\bottomrule
\end{tabular}
\label{tab:UterUS_UMD_TTA_results}
\end{table}

\begin{table}[t]
\centering
\caption{Segmentation metrics comparison of MedSAM2\_US\_Heart argumentation on Task08\_HepaticVessel dataset.}
\begin{tabular}{lccccc}
\toprule
        \textbf{Methods} & \textbf{mIoU} & \textbf{aIoU} & \textbf{mDice} & \textbf{aDice} & \textbf{HD95} \\ \midrule
        Baseline & 75.87 & 81.21 & 82.28 & 88.16 & 33.06 \\
        Gamma correction    & 75.74\textsubscript{\textcolor{red}{\hepGammaMIoU}} 
                            & 81.23\textsubscript{\textcolor{green}{\hepGammaAIoU}}
                            & 82.27\textsubscript{\textcolor{red}{\hepGammaMDice}}
                            & 88.38\textsubscript{\textcolor{green}{\hepGammaADice}}
                            & 32.83\textsubscript{\textcolor{green}{\hepGammaHD}} \\
        Contrast enhancement & 75.63\textsubscript{\textcolor{red}{\hepContrastMIoU}} 
                            & 81.32\textsubscript{\textcolor{green}{\hepContrastAIoU}}
                            & 82.19\textsubscript{\textcolor{red}{\hepContrastMDice}}
                            & 88.55\textsubscript{\textcolor{green}{\hepContrastADice}}
                            & 31.40\textsubscript{\textcolor{green}{\hepContrastHD}} \\
        Gaussian blur     & 78.12\textsubscript{\textcolor{green}{\hepBlurMIoU}} 
                          & 83.93\textsubscript{\textcolor{green}{\hepBlurAIoU}}
                          & 84.33\textsubscript{\textcolor{green}{\hepBlurMDice}}
                          & 90.12\textsubscript{\textcolor{green}{\hepBlurADice}}
                          & 38.65\textsubscript{\textcolor{red}{\hepBlurHD}}  \\
        Gaussian noise    & 79.06\textsubscript{\textcolor{green}{\hepNoiseMIoU}}
                          & 84.12\textsubscript{\textcolor{green}{\hepNoiseAIoU}}
                          & 84.93\textsubscript{\textcolor{green}{\hepNoiseMDice}} 
                          & 89.93\textsubscript{\textcolor{green}{\hepNoiseADice}} 
                          & 34.77\textsubscript{\textcolor{red}{\hepNoiseHD}}  \\
        \midrule
        Weighted Voting (0.6)    
                          & 77.47\textsubscript{\textcolor{green}{\hepVoteMIoU}}
                          & 83.15\textsubscript{\textcolor{green}{\hepVoteAIoU}}
                          & 83.70\textsubscript{\textcolor{green}{\hepVoteMDice}} 
                          & 89.60\textsubscript{\textcolor{green}{\hepVoteADice}} 
                          & 31.08\textsubscript{\textcolor{red}{\hepVoteHD}}  \\
\bottomrule
\end{tabular}
\end{table}

\begin{table}[!t]
\centering
\caption{Segmentation metrics comparison via single augmentation removal ablation on UterUS\cite{bones2024uterus} and UMD\cite{umd2024} datasets.}
\begin{tabular}{lccc|ccc}
\toprule
 & \multicolumn{3}{c}{\textbf{UterUS}} & \multicolumn{3}{c}{\textbf{UMD}} \\
\cmidrule(lr){2-4} 
\cmidrule(lr){5-7} 
 \textbf{Augmentations} & \textbf{IoU} & \textbf{Dice} & \textbf{HD95} & \textbf{IoU} & \textbf{Dice} & \textbf{HD95} \\
\midrule
Baseline & 81.23 & 89.26 & 33.61 & 83.68 & 87.91 & 37.16 \\
w/o Gamma correction  
         & 76.54\textsubscript{\textcolor{red}{\uterusRmGammaIoU}} 
         & 86.14\textsubscript{\textcolor{red}{\uterusRmGammaDice}} 
         & 25.08\textsubscript{\textcolor{green}{\uterusRmGammaHD}} 
         & 82.53\textsubscript{\textcolor{red}{\umdRmGammaIoU}}
         & 86.89\textsubscript{\textcolor{red}{\umdRmGammaDice}}
         & 34.20\textsubscript{\textcolor{green}{\umdRmGammaHD}}\\
w/o Contrast enhancement 
         & 76.64\textsubscript{\textcolor{red}{\uterusRmContrastIoU}} 
         & 86.18\textsubscript{\textcolor{red}{\uterusRmContrastDice}} 
         & 25.00\textsubscript{\textcolor{green}{\uterusRmContrastHD}}
         & 82.21\textsubscript{\textcolor{red}{\umdRmContrastIoU}}
         & 86.65\textsubscript{\textcolor{red}{\umdRmContrastDice}}
         & 33.89\textsubscript{\textcolor{green}{\umdRmContrastHD}}\\
w/o Gaussian blur 
         & 78.69\textsubscript{\textcolor{red}{\uterusRmBlurIoU}} 
         & 87.58\textsubscript{\textcolor{red}{\uterusRmBlurDice}} 
         & 30.07\textsubscript{\textcolor{green}{\uterusRmBlurHD}}
         & 81.01\textsubscript{\textcolor{red}{\umdRmBlurIoU}}
         & 85.30\textsubscript{\textcolor{red}{\umdRmBlurDice}}
         & 33.17\textsubscript{\textcolor{green}{\umdRmBlurHD}}\\
w/o Gaussian noise 
         & 77.01\textsubscript{\textcolor{red}{\uterusRmNoiseIoU}} 
         & 86.46\textsubscript{\textcolor{red}{\uterusRmNoiseDice}} 
         & 25.97\textsubscript{\textcolor{green}{\uterusRmNoiseHD}}
         & 77.73\textsubscript{\textcolor{red}{\umdRmNoiseIoU}}
         & 82.94\textsubscript{\textcolor{red}{\umdRmNoiseDice}}
         & 28.56\textsubscript{\textcolor{green}{\umdRmNoiseHD}}\\
\midrule
Weighted Voting (0.6)
         & 81.65\textsubscript{\textcolor{green}{\uterusVoteIoUDelta}} 
         & 89.60\textsubscript{\textcolor{green}{\uterusVoteDiceDelta}} 
         & 31.83\textsubscript{\textcolor{green}{\uterusVoteHDDelta}}
         & 84.17\textsubscript{\textcolor{green}{\umdVoteIoUDelta}}
         & 88.64\textsubscript{\textcolor{green}{\umdVoteDiceDelta}}
         & 34.43\textsubscript{\textcolor{green}{\umdVoteHDDelta}}\\
\bottomrule
\end{tabular}
\label{tab:Augmentation_ablation}
\end{table}

\subsection{Ablation Study}

We conducted ablation experiments to evaluate component contributions and parameter sensitivity in SegTTA.

\subsubsection{Augmentation Contribution Analysis}

To quantify the specific contribution of each augmentation component within the SegTTA framework, we conducted an ablation study by systematically removing one augmentation at a time. The detailed quantitative results are summarized in Table~\ref{tab:Augmentation_ablation}.

For the \textbf{UterUS dataset} (healthy uterus segmentation), we observed that intensity-based transformations are paramount. As shown in the table, the removal of Gamma correction and Contrast enhancement led to the most significant performance drops, with IoU decreasing by \uterusRmGammaIoU\% and \uterusRmContrastIoU\%, respectively. This distinct degradation suggests that the model relies heavily on robustness to global intensity shifts to accurately delineate the boundaries of large anatomical organs.

In contrast, the \textbf{UMD dataset} (uterine myoma detection) exhibited a different sensitivity profile. Here, the exclusion of Gaussian noise resulted in the sharpest decline in accuracy (IoU: \umdRmNoiseIoU\%, Dice: \umdRmNoiseDice\%). This indicates that noise-related augmentations are essential for distinguishing small, subtle lesions from the heterogeneous tissue background.

To quantify the specific contribution of each augmentation component, we conducted an ablation study by removing one augmentation at a time. The detailed quantitative comparisons are presented in Table~\ref{tab:Augmentation_ablation}.

For the \textbf{UterUS dataset} (large structure), intensity-based augmentations proved most critical. Specifically, removing Gamma correction and Contrast enhancement resulted in the largest performance drops, with IoU decreasing by \uterusRmGammaIoU\% and \uterusRmContrastIoU\%, respectively. 
In contrast, the \textbf{UMD dataset} (small lesions) showed greater sensitivity to noise. The removal of Gaussian noise caused the sharpest decline in accuracy (IoU: \umdRmNoiseIoU\%, Dice: \umdRmNoiseDice\%).

These findings highlight a critical insight: augmentation strategies must align with anatomical characteristics. While large structures benefit from global intensity variations, small targets require noise resilience to enhance local contrast. Furthermore, the consistent improvement in HD95 across all ablation settings confirms that the ensemble voting mechanism effectively refines boundary precision, regardless of the specific augmentation removed.

\begin{table}[t]
\centering
\caption{Segmentation metrics comparison with voting thresholds on UterUS\cite{bones2024uterus} and UMD\cite{umd2024} datasets}
\begin{tabular}{lccc|ccc}
\toprule
 & \multicolumn{3}{c}{\textbf{UterUS}} & \multicolumn{3}{c}{\textbf{UMD}} \\
\cmidrule(lr){2-4} 
\cmidrule(lr){5-7} 
 \textbf{Thresholds} & \textbf{IoU} & \textbf{Dice} & \textbf{HD95} & \textbf{IoU} & \textbf{Dice} & \textbf{HD95} \\
\midrule
0.6  & 81.65 & 89.60 & 31.83 & 84.17 & 88.64 & 34.43 \\
0.3  
         & 86.80\textsubscript{\textcolor{green}{\uterusLowIoU}} 
         & 92.73\textsubscript{\textcolor{green}{\uterusLowDice}} 
         & 38.86\textsubscript{\textcolor{red}{\uterusLowHD}} 
         & 90.93\textsubscript{\textcolor{green}{\umdLowIoU}}
         & 94.34\textsubscript{\textcolor{green}{\umdLowDice}}
         & 52.55\textsubscript{\textcolor{red}{\umdLowHD}}\\
0.9 
         & 76.13\textsubscript{\textcolor{red}{\uterusHighIoU}} 
         & 85.84\textsubscript{\textcolor{red}{\uterusHighDice}} 
         & 24.88\textsubscript{\textcolor{green}{\uterusHighHD}}
         & 74.29\textsubscript{\textcolor{red}{\umdHighIoU}}
         & 79.60\textsubscript{\textcolor{red}{\umdHighDice}}
         & 25.10\textsubscript{\textcolor{green}{\umdHighHD}}\\
\bottomrule
\end{tabular}
\label{tab:threshold_ablation}
\end{table}

\begin{figure}[t]
    \centering
    \includegraphics[width=1\textwidth]{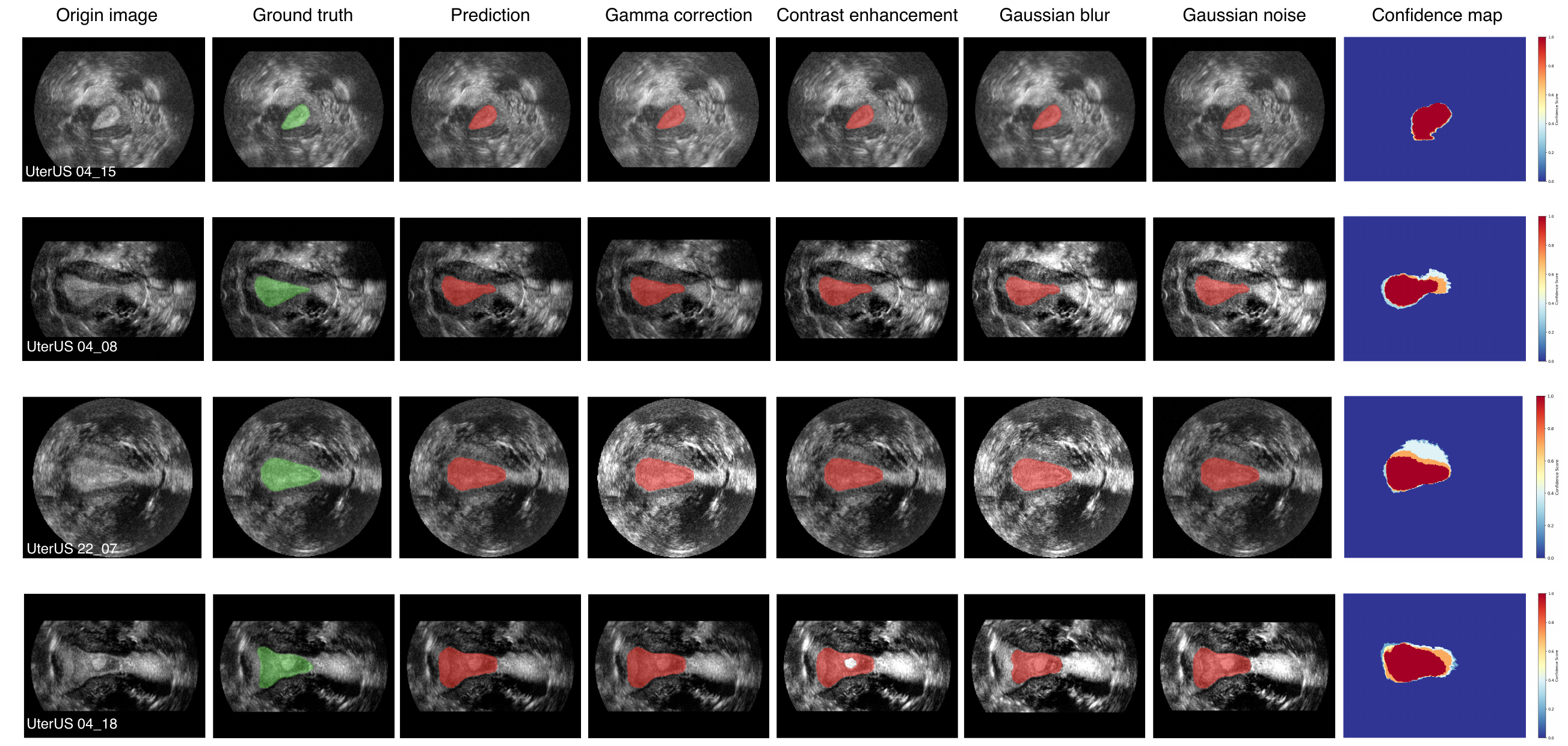}
    \caption{Qualitative segmentation results on the UterUS dataset. SegTTA demonstrates improved boundary delineation for the endometrial cavity compared to the baseline.}
    \label{fig:uterus_qualitative}
\end{figure}

\subsubsection{Voting Threshold Sensitivity}

Table~\ref{tab:threshold_ablation} examines voting threshold impact. 
Lower threshold (0.3) improved IoU/Dice (UterUS: $\uterusLowIoU\%$/$\uterusLowDice\%$, 
UMD: $\umdLowIoU\%$/$\umdLowDice\%$) but degraded HD95 
($\uterusLowHD$\,mm/$\umdLowHD$\,mm). 
The HD95 degradation was more pronounced for myoma segmentation, reflecting the challenge of precise boundary delineation for small structures. 
Higher threshold (0.9) enhanced boundary accuracy (HD95: $\uterusHighHD$\,mm/$\umdHighHD$\,mm) 
while reducing IoU ($\uterusHighIoU\%$/$\umdHighIoU\%$), with myoma segmentation showing greater sensitivity due to its smaller target volume.

The default threshold (0.6) provides optimal balance for both anatomical scales, though clinical applications may benefit from task-specific tuning: lower thresholds for complete organ coverage, higher thresholds for precise lesion boundaries.

\section{Qualitative Evaluation}

We present a visual comparison between the baseline MedSAM2 and our proposed SegTTA framework across three datasets. Figure~\ref{fig:uterus_qualitative} illustrates the segmentation results on the UterUS dataset. The baseline model exhibits jagged edges and ambiguity in low-contrast regions. in contrast, SegTTA produces smoother and more accurate boundaries for the endometrial cavity, validating the effectiveness of intensity-based augmentations for large anatomical structures.

\begin{figure}[!h]
    \centering
    \includegraphics[width=1\textwidth]{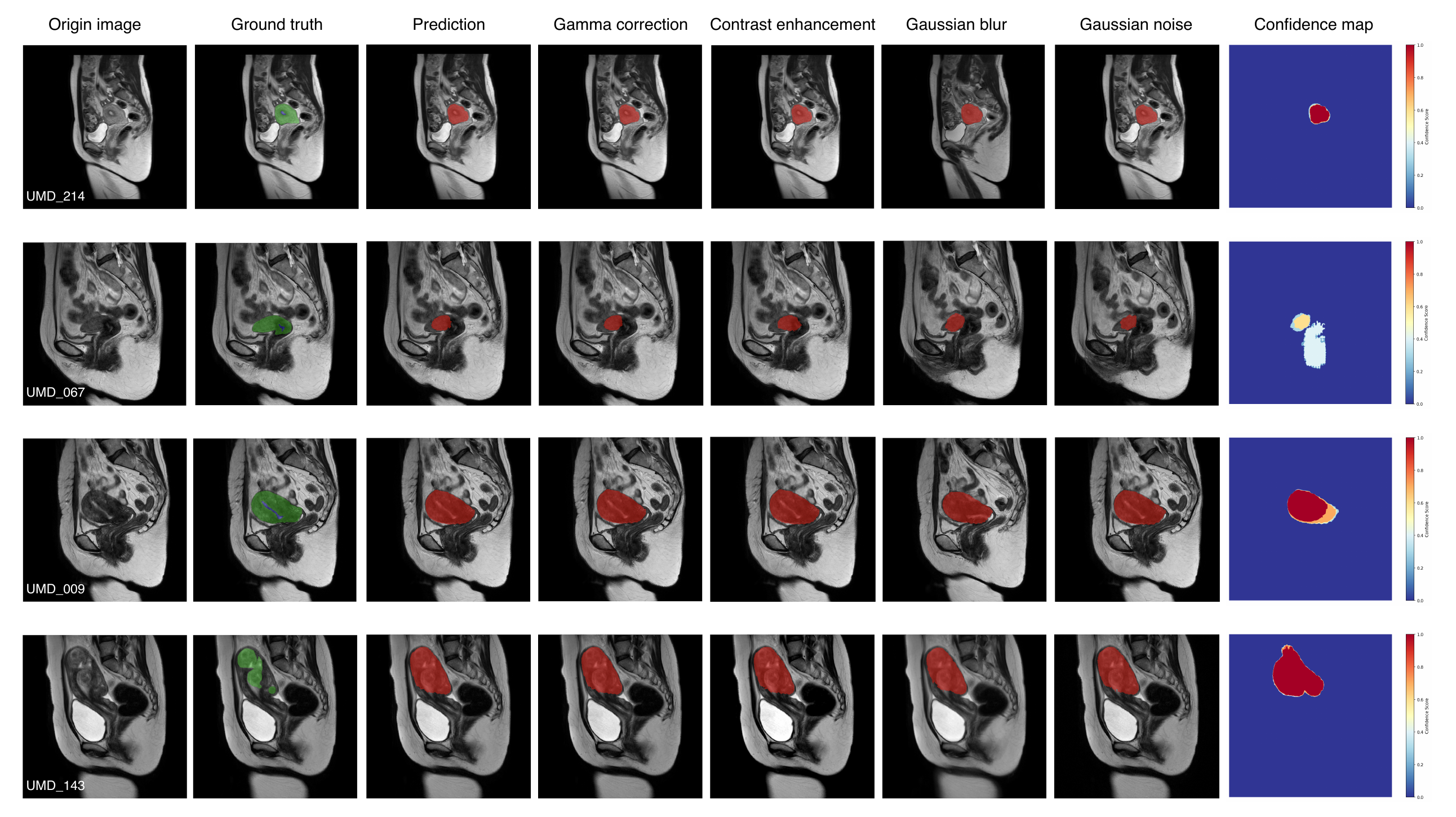}
    \caption{Visual comparison on the UMD dataset for uterine myoma detection. The proposed method effectively reduces noise interference and accurately identifies small lesion targets.}
    \label{fig:umd_qualitative}
\end{figure}

Figure~\ref{fig:umd_qualitative} displays the performance on the UMD dataset for uterine myoma detection. The baseline frequently misses small lesions or generates false positives due to background noise. SegTTA effectively suppresses these artifacts and improves the recall of small myoma targets, which aligns with our finding that noise augmentations are critical for small lesion segmentation.

\begin{figure}[!h]
    \centering
    \includegraphics[width=1\textwidth]{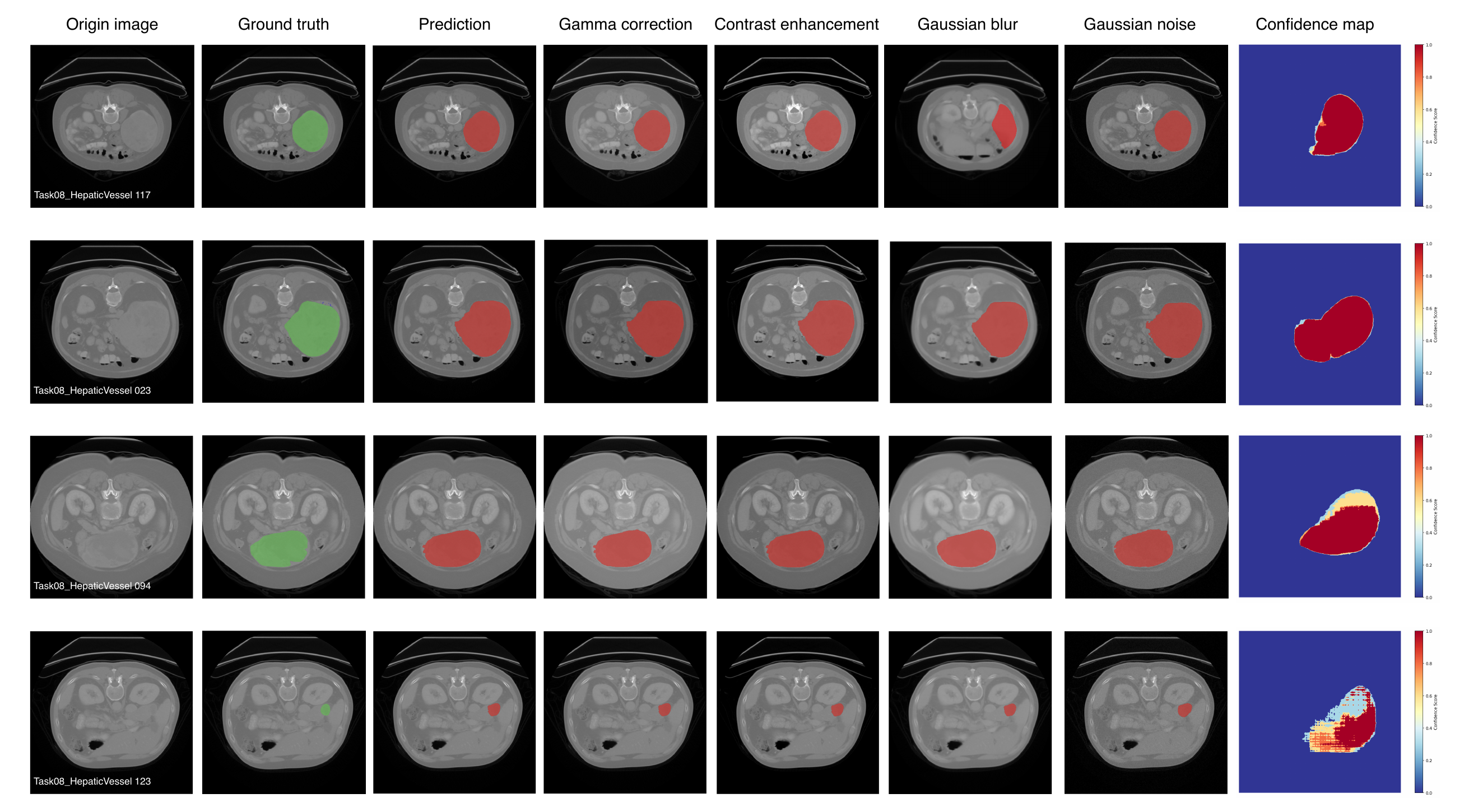}
    \caption{Segmentation results on the Task08 HepaticVessel dataset. SegTTA enhances the structural continuity of hepatic vessels and tumors in multi-class segmentation tasks.}
    \label{fig:hepatic_qualitative}
\end{figure}

Results for the multi-class HepaticVessel dataset are shown in Figure~\ref{fig:hepatic_qualitative}. SegTTA demonstrates superior capability in maintaining the structural continuity of hepatic vessels and clearly delineating tumors from surrounding tissues. The ensemble approach mitigates the instability of single-model predictions, resulting in more coherent multi-class segmentation maps.

\section{Limitation and Future Work}
While SegTTA demonstrates robust performance in a training-free manner, it inherently increases computational cost and inference latency due to the necessity of processing multiple augmented views and aggregating them through the voting mechanism. Currently, the framework relies on a fixed set of four augmentations  and manually adjusted voting thresholds, which, although effective across tested datasets, may not dynamically adapt to the unique noise characteristics of every individual clinical case. Future work will focus on addressing these efficiency bottlenecks by exploring adaptive test-time policies that automatically select the most relevant augmentations based on input uncertainty, thereby optimizing the trade-off between segmentation accuracy and real-time deployment feasibility.

\section{Conclusion}

SegTTA demonstrates that targeted test-time augmentation can significantly enhance medical image segmentation across diverse anatomical structures. Our framework achieved consistent improvements on three distinct segmentation tasks: healthy uterus (81.65\% IoU), uterine myoma (84.17\% IoU), and multi-class hepatic structures (77.47\% mIoU). The success across these varied targets, from large organs to small lesions to multi-class structures, validates the framework's versatility. Ablation studies revealed important insights about augmentation strategies: intensity-based transformations prove crucial for large organ boundaries, while noise augmentations excel at enhancing local contrast for small lesion detection. The adjustable voting threshold emerged as a powerful tool for clinical customization, allowing practitioners to prioritize either segmentation completeness or boundary precision based on diagnostic requirements. These findings indicate that effective TTA must consider anatomical characteristics rather than applying uniform strategies, providing a practical path to improve existing models without costly retraining.

\clearpage
\bibliographystyle{unsrt}  
\bibliography{references}

\end{document}